\documentclass{article} % For LaTeX2e
\usepackage{iclr2027_conference,times}

\usepackage{amsmath,amsfonts,bm}

\def\eqref#1{equation~\ref{#1}}
\def\1{\bm{1}}

\DeclareMathAlphabet{\mathsfit}{\encodingdefault}{\sfdefault}{m}{sl}
\SetMathAlphabet{\mathsfit}{bold}{\encodingdefault}{\sfdefault}{bx}{n}

\usepackage{hyperref}
\usepackage{url}
\usepackage{booktabs}
\usepackage{graphicx}
\usepackage{multirow}
\usepackage{etoolbox}
\usepackage{xcolor}
\definecolor{urlblue}{RGB}{12,92,171}

\newcommand{\hficon}[1][1.05em]{%
  \raisebox{-0.2\height}{\includegraphics[height=#1]{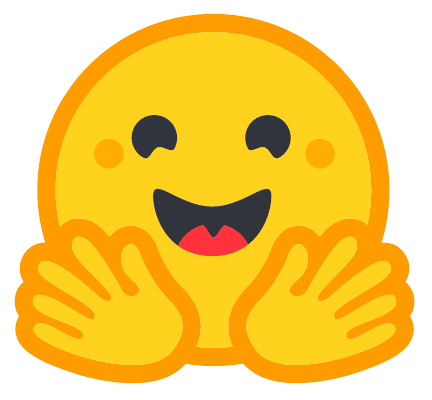}}%
}

\makeatletter

\patchcmd{\@maketitle}
    {\@title\par}
    {\@title\par\vskip 12pt}
    {}{}

\patchcmd{\@maketitle}
    {\vskip 0.3in minus 0.1in}
    {\vskip 10pt}
    {}{}

\makeatother

\title{GIFT-Eval-CTX: A Benchmark For Contextual Forecasting Model Evaluation}

\author{%
    \begin{minipage}{\dimexpr\textwidth-2\tabcolsep\relax}
        \normalfont
        \centering
        \textbf{Xu Liu, Hanshu Yan\thanks{Corresponding authors. Emails: hanshu.yan@salesforce.com, junnan.li@salesforce.com}, Taha Aksu, Quang Pham,}\\[0.3em]
        \textbf{Zhongzheng Qiao, Jun Hao Liew, Silvio Savarese, Doyen Sahoo, Junnan Li\footnotemark[2]}\\[0.7em]
        \textit{Moirai Team, Salesforce AI Research}\\[0.7em]
        {%
            \hypersetup{pdfborder={0 0 0}}%
            \urlstyle{same}%
            \color{urlblue}%
            \hficon\, \url{https://huggingface.co/spaces/Salesforce/GIFT-Eval-CTX}%
        }
    \end{minipage}%
}

\iclrfinalcopy % Uncomment for camera-ready version, but NOT for submission.
\begin{document}

\vspace*{-26pt}
\maketitle

\begin{abstract}
Contextual forecasting seeks to improve time series forecasting by incorporating forward-looking information, such as planned events, external conditions, and expert expectations. However, reliable evaluation remains difficult: existing benchmarks often have limited domain coverage, lack standardized history--future splits, or provide text that is not guaranteed to inform the future. We introduce \textbf{GIFT-Eval-CTX}, a benchmark that isolates whether forecasting methods can translate verified future information into accurate numerical trajectories. Built upon GIFT-Eval, GIFT-Eval-CTX contains 10,000 instances across 20 datasets. Each instance provides structured effects derived from the realized future: Macro effects describe horizon-wide behavior, while Micro effects specify localized changes. We construct these annotations with an agentic pipeline combining LLM-based discovery and deterministic code validation. We benchmark TSFMs, frontier LLMs, and tool-using agentic systems. The results show that structured effects reduce aggregate relative MASE by 40.7\% for GPT-5.6 Sol and 29.6\% for Claude Opus 4.8 relative to history-only prompting. The evaluated MoiraiAgent system achieves the best overall result. We further analyze the contributions of structured effects and agentic systems to performance, and identify opportunities for future research.
\end{abstract}

\section{Introduction}
\label{sec:introduction}

Time series foundation models (TSFMs) have become strong zero-shot forecasters~\citep{timesfm,chronos2,moirai2,toto2}, but their standard interfaces are designed primarily for numerical histories and predefined covariates. Consequently, they are not naturally equipped to incorporate open-ended forward-looking information, such as an energy operator expecting unusually high electricity demand during a heatwave, a transportation system facing a planned road closure, or a domain expert specifying how demand may evolve under a future scenario. Contextual forecasting addresses this limitation by integrating textual context with numerical history to guide forecasts~\citep{chatts,unitime,chattime,timeomni1}. This requires two complementary capabilities: understanding what diverse, domain-specific context implies about the future and translating those implications into accurate numerical trajectories.

Developing and comparing contextual forecasting systems requires reliable benchmarks that pair diverse time series with relevant forward-looking context under standardized evaluation protocols. However, existing contextual forecasting benchmarks fall short of these requirements in several ways~\citep{fidelts,timemqa,finmultitime,mmtsbench,fintexts}. First, collecting aligned text and time series data is difficult, resulting in limited domain diversity. Second, many datasets lack standardized history–future splits (e.g., providing only several years of financial news and related stock prices). Researchers must construct these splits themselves, reducing comparability across studies. Third, the accompanying text is not guaranteed to contain information about future events or conditions, making it unclear whether the benchmark truly evaluates a model's ability to use forward-looking context.

\begin{figure*}[t]
    \centering
    \includegraphics[width=\textwidth]{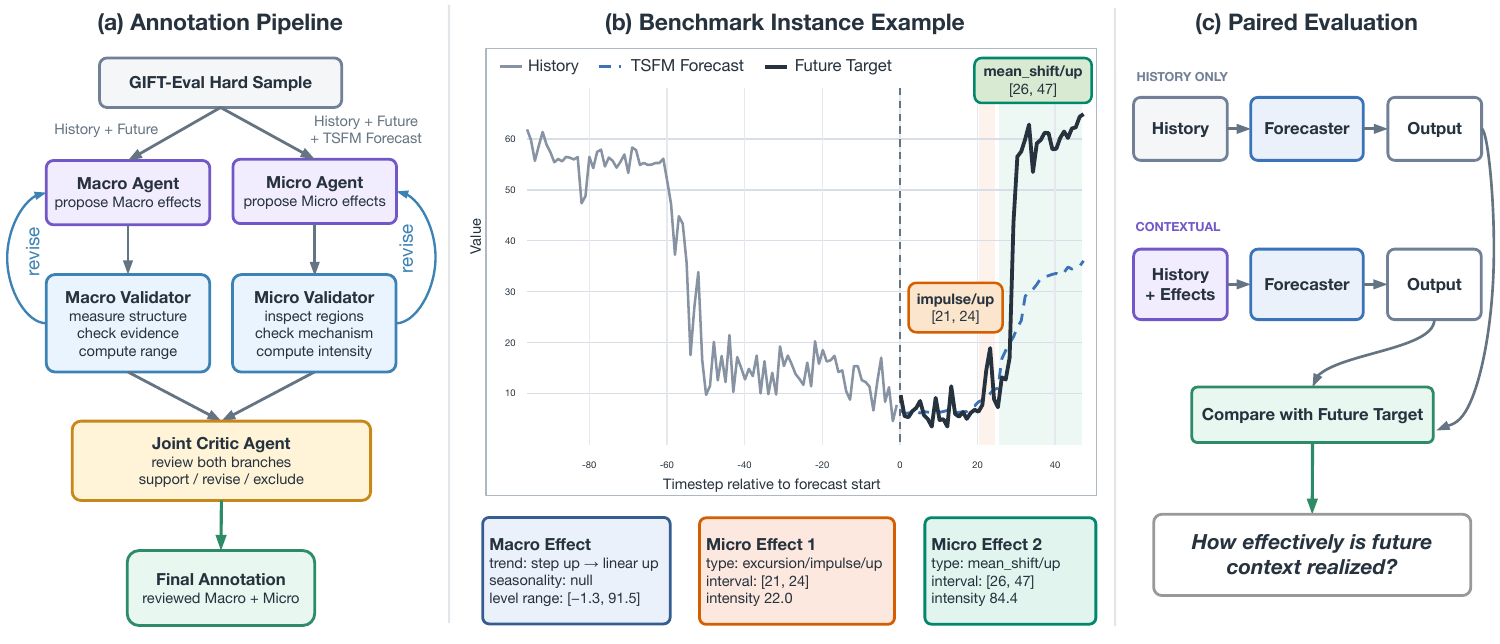}
    \vspace{-1em}
    \caption{Illustration of agentic annotation, an annotated instance, and paired evaluation.}
    \vspace{-1em}
    \label{fig:introduction}
\end{figure*}

In this work, rather than treating contextual forecasting as a single end-to-end task, we distinguish two capabilities it requires: \textit{contextual understanding} and \textit{numerical forecasting}. We treat contextual understanding as an upstream capability handled by frontier LLMs, whose knowledge and reasoning continue to improve rapidly, and focus on the evaluation of the second capability: generating accurate numerical forecasts from verified future information. To this end, we introduce \textbf{GIFT-Eval-CTX}, a contextual extension of the GIFT-Eval benchmark~\citep{gifteval}. GIFT-Eval offers broad domain diversity and standardized training--test and history--future splits, addressing the first two limitations above. To ensure that every context is informative about its corresponding future, we derive structured context from the realized future during benchmark construction, analogous to supplying a verified scenario or expert outlook. GIFT-Eval-CTX therefore does not test whether a model can extract or validate future information from raw sources, an upstream task that we believe can be handled by domain experts and increasingly capable frontier LLMs. Instead, it evaluates whether a model can convert provided future information into an accurate numerical trajectory.

To make the context portable across domains and less dependent on the language style of any particular LLM, we represent it as a compact key-value schema called \textbf{structured effects}, rather than as free-form text. Structured effects have two compositional levels. \textbf{Macro effects} describe horizon-wide trend, seasonality, and value range. \textbf{Micro effects} describe localized excursions, mean shifts, and seasonal changes, together with their temporal boundaries and scale. We then propose an agentic labeling pipeline to generate these structured effects. Concretely, we first identify GIFT-Eval test windows on which state-of-the-art TSFMs cannot solve easily, i.e., where future context is needed. Then a Macro Agent and a Micro Agent inspect the numerical evidence independently: the former summarizes global future structure, while the latter identifies localized mechanisms that are not adequately captured by the TSFMs' forecast. Deterministic validators enforce the schema, verify numerical evidence, and compute quantitative fields, after which a Joint Critic Agent reviews all proposed labels and commits a final annotation.

We benchmark three classes of forecasting methods on GIFT-Eval-CTX: TSFMs, frontier LLMs, and agentic systems that augment LLMs with tool use. The results show that LLM forecasts improve substantially when given structured effects: aggregate relative MASE decreases by 40.7\% for GPT-5.6 Sol and 29.6\% for Claude Opus 4.8 relative to history-only prompting. These improvements demonstrate that the structured effects are informative and actionable. Moreover, MoiraiAgent, which combines a frontier LLM with code-execution tools and a TSFM forecasting service, achieves the best overall performance. We also analyze how Macro and Micro effects contribute across forecast horizons, Micro families, and regions inside and outside labeled Micro intervals. Finally, we examine the sources of MoiraiAgent's gains, showing how its forecasting and code-execution tools improve performance, and identify limitations of code-based forecast editing.

\section{Related Work}
\label{sec:related-work}

\subsection{Multimodal Time Series Benchmarks}

Multimodal time series benchmarks have expanded time series evaluation beyond point prediction. One line of work focuses on understanding and question answering across a diverse collection of tasks~\citep{llmstruggle, temporalbench}. MTBench~\citep{mtbench} and Time-MQA~\citep{timemqa} pair numerical sequences with textual context and questions that require temporal or cross-modal reasoning. MMTS-Bench~\citep{mmtsbench} and TSRBench~\citep{tsrbench} broaden the scope to tasks such as pattern perception, feature analysis, and decision making. BEDTime~\citep{bedtime} instead evaluates the ability to describe and distinguish time series patterns in text.

A second line of work concentrates more directly on multimodal forecasting. Context is Key~\citep{cik} pairs numerical series with manually constructed, and sometimes intervention-aligned, synthetic textual contexts. Time-MMD~\citep{timemmd} aligns time series with real textual information across domains, and MoTime~\citep{motime} augments forecasting datasets with text, metadata, and images.  FinTexTS~\citep{fintexts} constructs text--time series pairs for financial forecasting through multi-level news matching. FinMultiTime~\citep{finmultitime} further extends financial multimodality by incorporating financial tables and K-line charts. However, these existing benchmarks often have limited domain coverage and lack standardized history--future splits. Moreover, their text is not guaranteed to provide future-relevant information. GIFT-Eval-CTX complements these benchmarks with broad domain coverage, standardized history--future splits, and structured context guaranteed to describe the realized future. It thereby isolates a specific downstream capability: translating provided future information into an accurate numerical trajectory.

\section{GIFT-Eval-CTX}
\label{sec:gift-eval-ctx}

\subsection{Task Definition}
\label{sec:task-definition}

Let $\mathbf{x}_{1:T} \in \mathbb{R}^{T}$ denote an observed time series history and $\mathbf{y}_{1:H} \in \mathbb{R}^{H}$ its future target over a horizon of $H$ steps. In contextual forecasting, the model additionally receives structured future context $\mathbf{s}$ describing the global behavior and localized effects within $\mathbf{y}_{1:H}$. Given $(\mathbf{x}_{1:T}, H, \mathbf{s})$, a contextual forecaster $F$ produces $\hat{\mathbf{y}}_{1:H} = F(\mathbf{x}_{1:T}, H, \mathbf{s})$. GIFT-Eval-CTX supports a paired evaluation of the same forecasting model under two information conditions. In the \emph{history-only} condition, the model receives only $(\mathbf{x}_{1:T}, H)$. In the \emph{contextual} condition, it additionally receives $\mathbf{s}$. Comparing the two conditions allows us to assess how effectively a model uses the supplied future context.

% This use of future ground truth is deliberate: $\mathbf{s}$ is privileged information that stands in for a verified future-facing signal, such as an expert outlook or an anticipated scenario. The benchmark therefore does not measure whether a model can infer this information from the historical series, nor does it claim that the information would be available in conventional history-only forecasting.

% More broadly, a real contextual forecasting system may first extract a structured representation from raw sources such as documents, planned actions, domain experts, or external events, and then translate that representation into a numerical trajectory. GIFT-Eval-CTX focuses on the second stage. Acquiring context from raw sources and assessing the factual or causal validity of that context remain outside the present benchmark; the task is to realize already-provided structured future information accurately in time and value. And our benchmark results indicate that this task still has significant room for improvement.

\subsection{Time Series Corpus}
\label{sec:time-series-corpus}

\paragraph{Data source.}
We construct GIFT-Eval-CTX from the held-out test regions of GIFT-Eval. Each multivariate series is flattened into univariate targets, and no window is cut from a training or validation region. We consider forecast horizons of at least 24 steps, since shorter futures rarely contain meaningful effects. To limit token costs for long histories, we cap the history length at $10H$, $5H$, and $3H$ for short-, medium-, and long-horizon settings, respectively.

\paragraph{Difficulty filtering.}
An important design principle of GIFT-Eval-CTX is to focus on test windows that TSFMs cannot solve easily. If a TSFM already forecasts the future accurately from history alone, additional context leaves little room for meaningful improvement. To identify non-trivial windows, we forecast every candidate with top-performing TSFMs on the GIFT-Eval leaderboard, and define the range-normalized error $\operatorname{rMAE}(\hat{\mathbf{y}}, \mathbf{y}) = \frac{H^{-1}\sum_{t=1}^{H}|\hat{y}_t-y_t|}{\max_t y_t-\min_t y_t}$. We retain a window only when all evaluated models obtain an rMAE larger than a threshold. Normalizing by the future range makes the threshold comparable across datasets and windows with different scales.

\subsection{Structured Effects Schema}
\label{sec:schema}

Free-form descriptions confound the information conveyed by a context with its wording, domain vocabulary, and level of detail. We instead represent future context through \emph{structured effects}, a domain-agnostic key-value schema designed to be portable across models. Structured effects decompose the future into a single Macro description and a set of Micro effects.

\paragraph{Macro effects.}
Macro describes the complete forecast horizon through three fields:
\begin{itemize}
    \item \texttt{trend}: one or two chronological segments describing the horizon-wide baseline movement. Each segment specifies a \texttt{direction} that describes the movement of the center line within the segment, a \texttt{shape} that describes the curvature of the trend, and a \texttt{duration} that gives its fraction of the horizon. Segment durations sum to one.
    \item \texttt{seasonality}: the dominant oscillatory structure of the future, or null when no meaningful oscillation is present. When present, it specifies a \texttt{period} in time steps and a \texttt{waveform} that describes the cycle shape.
    \item \texttt{level\_range}: the min and max future values in history-normalized coordinates.
\end{itemize}

\paragraph{Micro effects.}
Each Micro effect occupies an interval $[a,b]$ in the forecast horizon and describes a localized mechanism that is not adequately captured by the TSFMs' forecast. Multiple Micro effects may coexist or overlap when they describe different mechanisms. Micro has the following fields:
\begin{itemize}
    \item \texttt{family}: \texttt{excursion} (a temporary deviation followed by recovery), \texttt{mean\_shift} (a persistent transition from the nearest preceding stable level), or \texttt{seasonal\_change} (a change in a periodic pattern relative to the nearest preceding stable periodic region).
    \item \texttt{subtype}: \texttt{impulse}, \texttt{sustained}, or \texttt{recurrent} for excursions; null for mean shifts; \texttt{amplitude}, \texttt{phase}, \texttt{period}, \texttt{onset}, or \texttt{disappearance} for seasonal changes.
    \item \texttt{direction}: the direction relative to the local background or preceding reference regime. Its possible values depend on the family and subtype.
    \item \texttt{start}, \texttt{end}: the zero-based inclusive boundaries of the effect within the forecast horizon.
    \item \texttt{intensity}: a value computed deterministically from the raw series to quantify the effect. Its interpretation is family-specific.
    \item \texttt{amplitude\_trend}: how the amplitude evolves within the effect: \texttt{increase}, \texttt{decrease}, \texttt{stable}, or \texttt{variable}.
    \item \texttt{envelope\_change}: how the amplitude envelope evolves within the effect: \texttt{expanding}, \texttt{contracting}, \texttt{stable}, or \texttt{variable}.
\end{itemize}

To make quantitative fields comparable across series, Macro \texttt{level\_range} and Micro \texttt{intensity} values use a shared history-normalized scale. Let $x_{\min}$ and $x_{\max}$ be the minimum and maximum finite history values. A normalized level $v$ is represented as $\ell(v)=100\frac{v-x_{\min}}{x_{\max}-x_{\min}}$. Values below 0 or above 100 are valid when the future falls outside the historical range. This normalization preserves relative scale while allowing one schema to cover datasets with different units and magnitudes. \textbf{Complete schemas and visualization are shown in Appendix \ref{app:macro-schema}, \ref{app:micro-schema} and \ref{app:micro-effect-examples}.}

\subsection{Structured Effects Generation Pipeline}
\label{sec:generation-pipeline}

\paragraph{Overview.}
We propose an agentic pipeline with an inspect--propose--validate--accept/revise workflow. As shown in Figure~\ref{fig:introduction}, the pipeline first runs independent Macro and Micro Discovery Agents, each combining LLM reasoning with deterministic code validation. The LLMs identify informative regions, determine effects' mechanism and boundaries, and submit proposals. Deterministic validation then checks the proposed type and type-specific criteria. If a proposal fails validation, the tool returns the numerical evidence so that the Agent can revise its proposal. Finally, a Joint Critic Agent reviews the outputs of both branches and filters out incorrect annotations. Moreover, a random subset of annotated samples is manually reviewed to assess annotation quality.

\paragraph{Macro Discovery Agent.}
The Macro Agent observes the complete history and future ground truth. It first examines the raw series and calls deterministic tools to measure trend, change-point, and periodic structure. Based on this evidence, it proposes a complete Macro object covering the forecast horizon. A Macro validator checks the schema and segment durations and computes the exact future range from the raw series. Measurements that disagree with the proposed trend or seasonality are returned as evidence for reconsideration. The Agent can then revise an invalid or poorly supported proposal before submitting its final Macro candidate.

\paragraph{Micro Discovery Agent.}
To focus Micro discovery on regions where TSFMs forecast poorly, we construct a reference forecast
for each sample by averaging the predictions of the best-performing TSFMs. The Micro Agent first calls a deterministic overview tool to locate candidate regions with large residuals between the reference forecast and the future ground truth. The Agent then calls inspection tools to read the raw values in each selected region and its surrounding context. From this evidence, it determines whether the region contains an independent effect and, if so, identifies its mechanism and complete boundaries. It then submits a proposal specifying the family, subtype, direction, interval, and any preceding reference region required by the mechanism.

Each proposal is submitted to a family-specific Micro validator through a tool call. The validator checks the schema, boundaries, and any required component related to the Micro effect. When a proposal fails these checks, the tool returns the relevant numerical evidence so that the Agent can inspect the region again and revise its interpretation. The Agent repeats this process to identify additional effects.

\paragraph{Blind-first Joint Critic Agent.}
After both discovery branches finish, a Joint Critic Agent reviews their outputs. The Critic first inspects the raw history and future while all outputs of the discovery agents remain hidden. Only after forming an independent interpretation can it view the proposed Macro and Micro effects, their explanations, and the corresponding numerical evidence. This order reduces the risk that the Critic simply adopts the discovery Agents' interpretations.

The Critic assesses the Macro claims and decides whether each Micro candidate should be supported, excluded, or left unresolved. Supporting a Micro candidate requires inspection of its complete interval and any surrounding region on which the effect depends. The Critic can also correct a candidate's type or boundaries, after which the revised candidate is checked again by the validator. The final annotation contains the reviewed Macro and only the supported Micro effects; uncertain candidates are not promoted to the final annotation.

\section{Experiments}
\label{sec:experiments}

\subsection{Experimental Setup}
\label{sec:experimental-setup}

\paragraph{Dataset.}
The final GIFT-Eval-CTX benchmark contains 10,000 instances from 20 GIFT-Eval source datasets and 78 dataset--frequency--term configurations. It covers six sampling frequencies and seven distinct horizon lengths between 30 and 900 steps. Short horizons account for 6,559 instances, medium horizons for 2,135, and long horizons for 1,306. Detailed source-level counts are deferred to Appendix \ref{app:benchmark-statistics}. The benchmark contains 23,476 Micro effects. Excursions are the most common family, with 19,134 effects, followed by 3,117 seasonal changes and 1,225 mean shifts.

\paragraph{Evaluated methods.}
Our experiments cover three classes of forecasting methods. The TSFM baselines are Chronos-2~\citep{chronos2}, Toto-2~\citep{toto2}, Moirai-2~\citep{moirai2}, and TimesFM-3~\citep{timesfm}, each operating in the univariate setting using only the numerical history. GPT-5.6 Sol~\citep{gpt56sol} and Claude Opus 4.8~\citep{claudeopus48} serve as frontier-LLM baselines and are tested under paired history-only and contextual conditions. Finally, we evaluate a GPT-5.6 Sol-powered MoiraiAgent~\citep{moiraiagent} that receives the structured effects and can use TSFM forecasting (powered by Moirai-2) and code-execution tools.

\paragraph{Metric and aggregation.}
Following GIFT-Eval, we use mean absolute scaled error (MASE) as the evaluation metric, which normalizes forecast error by the in-sample error of a seasonal-naive predictor. For each of the 78 dataset--term groups, we first average the per-sample MASE of each method and divide it by the corresponding seasonal-naive group mean. We then take the geometric mean across groups to obtain the aggregate relative MASE. Lower values are better, and seasonal naive has a score of 1.0 by construction.

\subsection{Main Results}
\label{sec:main-results}

\paragraph{Frontier LLMs still underperform TSFMs in history-only forecasting.}
Table~\ref{tab:main-results} reports MASE for all evaluated methods. Despite advances in the general knowledge, frontier LLMs still face limitations in history-only time series forecasting, consistent with earlier findings on less capable LLMs~\citep{llmstruggle, llmuseful}. Given history only, both frontier LLMs underperform every native TSFM in terms of overall MASE. Both models outperform seasonal naive at the short horizon, but this advantage disappears as the horizon increases. At the medium and long horizons, GPT-5.6 Sol performs 7.1\% and 4.9\% worse than seasonal naive, whereas Claude Opus 4.8 degrades more sharply, with errors increasing by 32.5\% and 20.6\%, respectively. In contrast, all TSFMs consistently outperform seasonal naive across all horizons.

\paragraph{Structured effects substantially improve frontier LLM forecasts.}
Providing structured effects reduces overall MASE by 40.7\% for GPT-5.6 Sol and 29.6\% for Claude Opus 4.8, enabling both models to achieve performance comparable to or better than TSFMs. These gains indicate that frontier LLMs can translate structured information into more accurate numerical trajectories. For both models, the benefit diminishes as the forecast horizon grows, with the largest improvements occurring at the short horizon and progressively smaller gains thereafter. Structured effects therefore remain useful beyond short horizons, but direct realization becomes increasingly difficult.

\paragraph{MoiraiAgent achieves the best overall performance.}
MoiraiAgent not only achieves the lowest overall MASE, but also performs best at every horizon. Relative to the best history-only TSFM overall, MoiraiAgent reduces MASE by 36.3\%. Its advantage over GPT-5.6 Sol with the same structured effects grows from 4.6\% on short horizons to 14.3\% on medium horizons and 22.9\% on long horizons, suggesting that tool use becomes increasingly valuable as the forecast horizon grows.

\begin{table*}[t]
    \centering
    \caption{MASE on GIFT-Eval-CTX. ``Effects'' denotes the structured context supplied by the benchmark. The best/second-best result in each column is bold/underlined.}
    \vspace{0.5em}
    \label{tab:main-results}
    \resizebox{0.76\textwidth}{!}{%
    \begin{tabular}{llrrrr}
        \toprule
        Method & Input & Short & Medium & Long & Overall \\
        \midrule
        Seasonal naive & History & 1.000 & 1.000 & 1.000 & 1.000 \\
        Chronos-2 & History & 0.764 & 0.803 & 0.848 & 0.795 \\
        Toto-2 & History & 0.765 & 0.811 & 0.851 & 0.798 \\
        Moirai-2 & History & 0.740 & 0.798 & 0.852 & 0.782 \\
        TimesFM-3 & History & 0.750 & 0.795 & 0.823 & 0.780 \\
        \midrule
        Claude Opus 4.8 & History & 0.967 & 1.325 & 1.206 & 1.109 \\
        Claude Opus 4.8 & History + effects & 0.538 & 1.077 & 1.147 & 0.781 \\
        GPT-5.6 Sol & History & 0.851 & 1.071 & 1.049 & 0.953 \\
        GPT-5.6 Sol & History + effects & \underline{0.417} & \underline{0.727} &
        \underline{0.782} & \underline{0.565} \\
        \midrule
        MoiraiAgent & History + effects & \textbf{0.398} & \textbf{0.623} &
        \textbf{0.603} & \textbf{0.497} \\
        \bottomrule
    \end{tabular}%
    }
\end{table*}

\subsection{Analysis of Structured Effects}
\label{sec:effects-analysis}

\paragraph{Micro effects drive the largest gains, while Macro effects remain complementary.}
Table~\ref{tab:effects-ablation} compares history-only forecasting with three structured-effect conditions. Micro effects are the larger individual contributor for both methods, while the additional improvement from combining Macro and Micro effects shows that the two levels provide complementary information.

The horizon breakdown shows that Macro and Micro effects contribute differently. For both GPT-5.6 Sol and MoiraiAgent, the gains from Macro effects diminish as the forecast horizon grows, whereas Micro effects yield substantial improvements at all three horizon lengths. The reason is that Macro effects provide coarse, horizon-wide information. As the horizon grows, many distinct trajectories may satisfy the same global description, making these effects less informative for constructing a long-horizon forecast. Micro effects remain more actionable because they specify localized changes. Nevertheless, Macro effects remain useful in combination with Micro effects: the combined condition outperforms Micro-only forecasting for both methods at every horizon.

\paragraph{Mean shifts yield the largest gains, while seasonal changes remain challenging.}
To compare the influence of three Micro-effect families, we group samples according to whether they contain at least one effect of excursion, mean shift, or seasonal change. Samples containing effects from multiple families are assigned to each corresponding group. As shown in Table~\ref{tab:micro-family-contributions}, both methods exhibit the same ordering of gains across the Micro families. Mean shifts produce the largest gains, reducing MASE by around two thirds. Excursions yield substantial but smaller reductions of approximately 29--31\%, whereas seasonal changes improve performance by only 6--9\% overall. The horizon breakdown shows that the gains from mean shifts and excursions are consistent across forecast lengths. Seasonal changes are more difficult: on long horizons, GPT-5.6 Sol deteriorates from 1.070 to 1.119 when Micro effects are provided, whereas MoiraiAgent improves from 0.826 to 0.755. This pattern suggests that persistent level changes are the most readily translated into numerical trajectories, while changes to periodic structure require coordinated adjustments across repeated cycles. The contrasting long-horizon results further indicate that tool-supported forecasting is more reliable than direct generation for realizing such distributed changes.

\begin{table*}[t]
    \centering
    \caption{MASE under different structured-effect inputs. MoiraiAgent's history-only performance is the same as Moirai-2.}
    \vspace{0.5em}
    \label{tab:effects-ablation}
    \resizebox{0.75\textwidth}{!}{%
        \begin{tabular}{llcccc}
            \toprule
            Method & Input & Short & Medium & Long & Overall \\
            \midrule
            \multirow{4}{*}{GPT-5.6 Sol} & History only & 0.851 & 1.071 & 1.049 & 0.953 \\
            & History + Macro & 0.601 & 0.992 & 1.009 & 0.780 \\
            & History + Micro & 0.520 & 0.796 & 0.815 & 0.650 \\
            & History + Macro + Micro & \textbf{0.417} & \textbf{0.727} &
            \textbf{0.782} & \textbf{0.565} \\
            \midrule
            \multirow{4}{*}{MoiraiAgent} & History only & 0.740 & 0.798 & 0.852 & 0.782 \\
            & History + Macro & 0.605 & 0.802 & 0.858 & 0.711 \\
            & History + Micro & 0.483 & 0.628 & 0.609 & 0.549 \\
            & History + Macro + Micro & \textbf{0.398} & \textbf{0.623} &
            \textbf{0.603} & \textbf{0.497} \\
            \bottomrule
        \end{tabular}%
    }
    % \vspace{-1em}
\end{table*}

\begin{table*}[t]
    \centering
    \caption{Comparison of MASE between history-only forecasting and forecasting with Micro effects. Each Micro family contains samples with at least one effect from that family.}
    \vspace{0.5em}
    \label{tab:micro-family-contributions}
    \small
    \setlength{\tabcolsep}{5pt}
    \begin{tabular}{@{}ll*{4}{cc}@{}}
        \toprule
        \multirow{2}{*}{Method} & \multirow{2}{*}{Micro family}
        & \multicolumn{2}{c}{Short} & \multicolumn{2}{c}{Medium}
        & \multicolumn{2}{c}{Long} & \multicolumn{2}{c}{Overall} \\
        \cmidrule(lr){3-4}
        \cmidrule(lr){5-6}
        \cmidrule(lr){7-8}
        \cmidrule(lr){9-10}
        & & Hist. & +Micro & Hist. & +Micro & Hist. & +Micro & Hist. & +Micro \\
        \midrule
        \multirow{3}{*}{GPT-5.6 Sol} & Excursion & 0.826 & 0.540 & 1.056 & 0.780 & 1.031 & 0.739 & 0.930 & 0.642 \\
        & Mean shift & 0.810 & 0.240 & 0.995 & 0.446 & 0.982 & 0.304 & 0.890 & 0.294 \\
        & Seasonal change & 0.857 & 0.753 & 1.104 & 1.025 & 1.070 & 1.119 & 0.975 & 0.914 \\
        \midrule
        \multirow{3}{*}{MoiraiAgent} & Excursion & 0.708 & 0.498 & 0.811 & 0.628 & 0.862 & 0.584 & 0.770 & 0.550 \\
        & Mean shift & 0.699 & 0.235 & 0.864 & 0.405 & 0.900 & 0.300 & 0.780 & 0.283 \\
        & Seasonal change & 0.742 & 0.678 & 0.866 & 0.792 & 0.826 & 0.755 & 0.796 & 0.728 \\
        \bottomrule
    \end{tabular}
    % \vspace{-1em}
\end{table*}

\paragraph{Micro effects improve forecasts both inside and between labeled regions.}
A successful forecast should follow each local instruction and connect separate effect regions through coherent numerical trajectories. We additionally examine how Micro effects change predictions outside their labeled regions. Table~\ref{tab:micro-region-performance} therefore reports performance separately inside and outside the labeled Micro regions. Within the labeled regions, adding Micro effects reduces GPT-5.6 Sol's MASE by 50.9\% and MoiraiAgent produces a similar 49.3\% reduction, confirming that both methods can effectively use Micro effects to realize the specified local events.

\begin{figure*}[t]
    \centering
    \includegraphics[width=\textwidth]{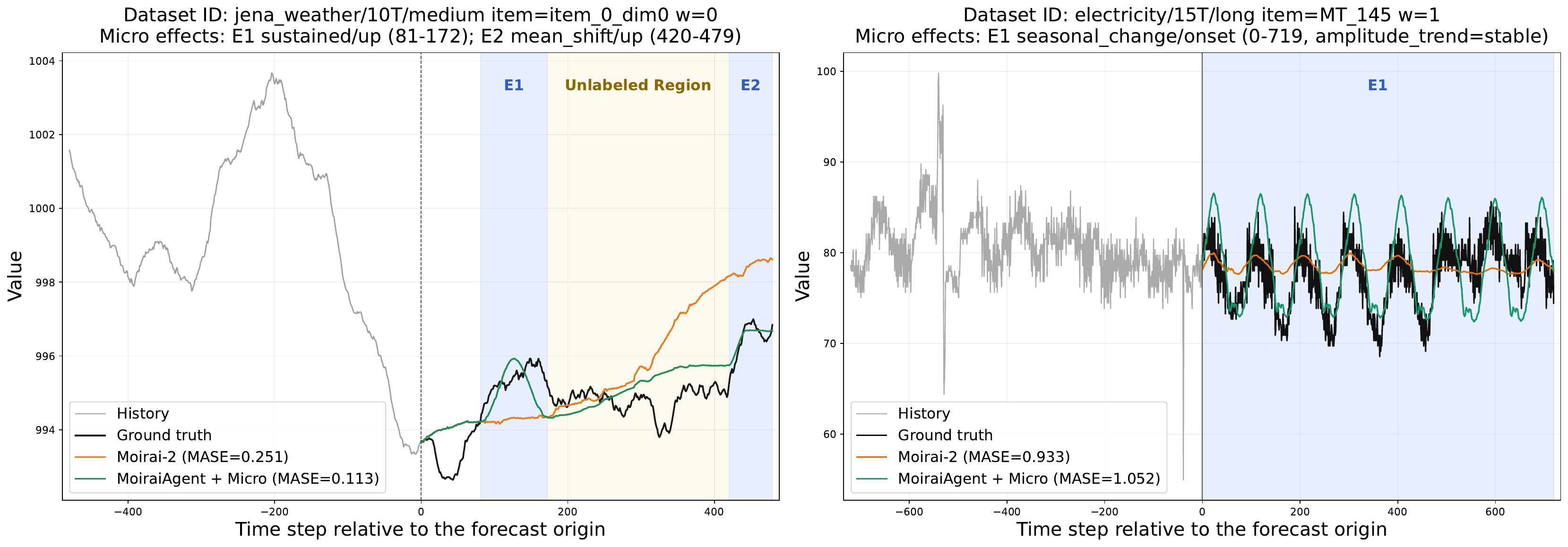}
    \caption{Qualitative examples of the benefits provided by Micro effects. Left: Micro effects improve the unlabeled trajectory between two labeled events. Right: Micro effects preserve meaningful periodic dynamics even when the sample-level MASE does not improve.}
    \label{fig:qualitative-effects}
\end{figure*}

Notably, both methods can propagate local effect information beyond the labeled regions. Outside the labeled regions, the overall MASE improves from 0.890 to 0.802 for GPT-5.6 Sol and from 0.687 to 0.652 for MoiraiAgent. Although smaller than the within-region gains, these improvements show that structured effects can positively influence the trajectory between local events. This also indicates that achieving strong performance on GIFT-Eval-CTX requires both realizing individual effects and connecting them through accurate and coherent numerical dynamics. The left panel of Figure~\ref{fig:qualitative-effects} illustrates this behavior. The sample contains an upward sustained excursion followed by an upward mean shift. Supplying these two Micro effects reduces MASE from 0.261 to 0.124 in the unlabeled interval between them, producing a more accurate transition between the two events.

\paragraph{Micro effects can preserve periodic structure even when MASE worsens.}
The right panel of Figure~\ref{fig:qualitative-effects} illustrates a complementary qualitative benefit that is not fully captured by the quantitative MASE metric. For this periodic electricity series, the history-only Moirai-2 forecast progressively attenuates the oscillations toward a nearly flat trajectory, exhibiting a common forecast-collapse behavior of TSFMs \citep{itstime}. Given a seasonal-onset effect with stable amplitude, MoiraiAgent instead preserves meaningful periodic dynamics throughout the forecast horizon, although its sample-level MASE increases from 0.933 to 1.052. Thus, this kind of Micro effects can prevent a periodic forecast from collapsing and preserve meaningful periodic structure even when this benefit is not reflected in the quantitative results.

\begin{table*}[t]
    \centering
    \caption{MASE inside and outside labeled Micro regions.}
    \vspace{0.5em}
    \label{tab:micro-region-performance}
    \resizebox{\textwidth}{!}{%
        \begin{tabular}{llcccccccc}
            \toprule
            \multirow{2}{*}{Method} & \multirow{2}{*}{Input} &
            \multicolumn{4}{c}{Inside Micro regions} &
            \multicolumn{4}{c}{Outside Micro regions} \\
            \cmidrule(lr){3-6}\cmidrule(lr){7-10}
            & & Short & Medium & Long & Overall & Short & Medium & Long & Overall \\
            \midrule
            \multirow{2}{*}{GPT-5.6 Sol} & History only &
            0.943 & 1.107 & 1.055 & 1.011 & 0.744 & 1.060 & 1.062 & 0.890 \\
            & History + Micro &
            0.411 & 0.650 & 0.542 & 0.496 & 0.638 & 0.918 & 1.101 & 0.802 \\
            \midrule
            \multirow{2}{*}{MoiraiAgent} & History only &
            0.838 & 0.893 & 0.939 & 0.876 & 0.610 & 0.729 & 0.820 & 0.687 \\
            & History + Micro &
            0.398 & 0.535 & 0.454 & 0.444 & 0.594 & 0.671 & 0.760 & 0.652 \\
            \bottomrule
        \end{tabular}%
    }
\end{table*}

\subsection{Further Analysis of MoiraiAgent}

MoiraiAgent achieves the best performance on GIFT-Eval-CTX. In this section, we investigate the sources of its strong performance and identify potential limitations to guide future improvements.

\paragraph{MoiraiAgent uses forecast tools to ensure exact-length outputs.}
Beyond accuracy, direct LLM forecasting exhibits a fundamental limitation in following the forecasting task's output-length requirement. As the horizon grows, LLMs may lose track of the required output length during generation, causing many medium- and long-horizon outputs whose lengths do not match the requested horizon. In such cases, we truncate overlong outputs and pad shorter ones to the required horizon. Table~\ref{tab:length-adjustment} shows that this issue is rare on short horizons but common on longer ones. GPT-5.6 Sol can only produce the requested length in 38.9\% of medium outputs and 40.5\% of long outputs. Claude Opus 4.8 shows even weaker length control, with corresponding rates of only 2.3\% and 1.2\%.

MoiraiAgent avoids this problem by first calling Moirai-2 to produce an exact-length numerical trajectory. The LLM uses code-execution tools to modify this existing forecast instead of constructing a long, numerically coherent forecast from scratch. Consequently, all MoiraiAgent forecasts have the requested length. This suggests that forecasting tools contribute more than a strong history-only baseline: they also provide an exact-length numerical object that the LLM can directly edit.

\paragraph{MoiraiAgent uses code execution to improve numerical constraint adherence.}
To evaluate whether a forecast numerically realizes the supplied structured effects, we measure its adherence to the specified level range and Micro-event intensities using a tolerance of five history-normalized points. Table~\ref{tab:length-adjustment} shows that the global level range is generally easier to follow than local event intensities. GPT-5.6 Sol follows the level range on 97.41\% of samples, but follows every specified intensity on only 81.57\% of samples. Claude Opus 4.8 is substantially less reliable on both constraints.

MoiraiAgent strengthens numerical constraint adherence through code execution and achieves the highest following rates. The code tool allows the Agent to calculate the numerical targets implied by the level range and event intensities, modify the relevant forecast regions, and verify the resulting values against these constraints. Together, these results show how the two tools address complementary limitations of direct LLM forecasting: the forecast tool provides an exact-length numerical starting point, while the code tool helps the Agent realize structured effects more precisely.

\paragraph{Limitations of MoiraiAgent.}
Code execution improves numerical constraint adherence, but it can also simplify the local dynamics of the forecast. MoiraiAgent often realizes an effect through deterministic operations such as level shifts, rescaling, and interpolation. These operations can place an event at the requested location and intensity, yet produce long constant segments or overly smooth linear transitions that do not capture the fine-grained temporal variation of time series.

\begin{figure*}[t]
    \centering
    \includegraphics[width=\textwidth]{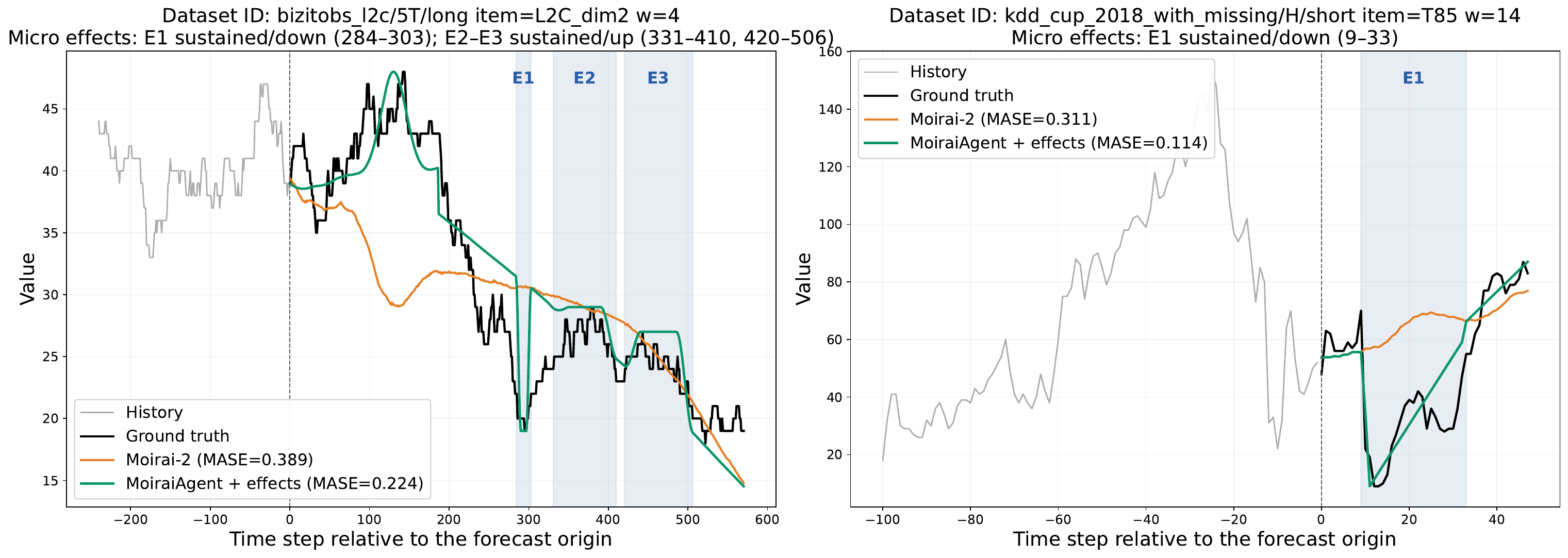}
    \caption{Qualitative limitations of MoiraiAgent. Left: MoiraiAgent replaces local fluctuations with extended plateaus in the BizITObs series. Right: it approximates the recovery from a sustained downward excursion with a nearly linear trajectory in the KDD Cup series. In both examples, sample-level MASE improves despite the loss of fine-grained temporal dynamics.}
    \vspace{-1em}
    \label{fig:agent-limitations}
\end{figure*}

Figure~\ref{fig:agent-limitations} shows two examples in which code-based editing fails to preserve realistic local dynamics. In the BizITObs series, MoiraiAgent reduces the sample-level MASE from 0.389 to 0.224, but represents the second and third sustained excursions with extended plateaus that suppress their local fluctuations. In the KDD Cup series, the MASE also improves, while the recovery from the sustained downward excursion is approximated by an almost perfectly straight line. Thus, future agentic forecasters should couple numerical constraint enforcement with more flexible mechanisms for preserving temporal variation during forecast editing. Another promising direction is to jointly train multimodal models on time series and structured effects, allowing them to represent and realize effect information natively within the forecasting model.

\begin{table*}[t]
    \centering
    \caption{Analysis of output-length compliance and numerical constraint adherence. T/P denotes the truncation and padding rates, respectively. The final two columns report sample-level adherence to the specified level range and event intensities using a tolerance of five history-normalized points.}
    \vspace{0.5em}
    \label{tab:length-adjustment}
    \resizebox{\columnwidth}{!}{%
    \begin{tabular}{lccccc}
        \toprule
        Method & Short (T/P) & Medium (T/P) & Long (T/P) & Follow level range &
        Follow intensity \\
        \midrule
        Claude Opus 4.8 & 0.0\% / 0.2\% & 48.4\% / 49.3\% & 34.0\% / 64.8\% & 59.83\% & 40.79\% \\
        GPT-5.6 Sol & 0.0\% / 0.1\% & 54.0\% / 7.1\% & 51.8\% / 7.7\% & 97.41\% & 81.57\% \\
        MoiraiAgent & 0.0\% / 0.0\% & 0.0\% / 0.0\% & 0.0\% / 0.0\% & 99.98\% & 92.73\% \\
        \bottomrule
    \end{tabular}%
    }
\end{table*}

\section{Conclusion}

We introduced GIFT-Eval-CTX, a benchmark that isolates the numerical realization of verified future information in time series forecasting. The benchmark represents context as compositional Macro and Micro effects and uses an agentic pipeline with deterministic validation to construct annotations across diverse datasets. Experiments show that structured effects substantially improve frontier LLM forecasts, while MoiraiAgent achieved the best performance by combining LLM reasoning with forecasting and code-execution tools. Our analyses also show that success requires both realizing the specified effects and maintaining coherent numerical dynamics throughout the forecast. These findings motivate two directions: building stronger agentic forecasters, or jointly training models on time series and structured effects for native effect representation and realization.

\subsection*{AI use statement}

In this work, we used generative AI tools to generate the structured-effect annotations, assist in implementing the code for baseline evaluations, and support the analysis of experimental results. We did not use generative AI to develop theoretical models, formulate mathematical claims, or produce proofs. Additionally, generative AI assisted with manuscript drafting and language polishing. The authors manually inspected annotation quality, executed and tested the code, verified the experimental outcomes, and reviewed the manuscript. We take responsibility for the final content of this work, including all text, claims, code, and artifacts produced with the aid of generative AI.

\bibliography{iclr2027_conference}

@inproceedings{moirai,
  title={Unified training of universal time series forecasting transformers},
  author={Woo, Gerald and Liu, Chenghao and Kumar, Akshat and Xiong, Caiming and Savarese, Silvio and Sahoo, Doyen},
  booktitle={International Conference on Machine Learning},
  year={2024}
}

@article{moirai2,
  title={Moirai 2.0: When less is more for time series forecasting},
  author={Liu, Chenghao and Aksu, Taha and Liu, Juncheng and Liu, Xu and Yan, Hanshu and Pham, Quang and Savarese, Silvio and Sahoo, Doyen and Xiong, Caiming and Li, Junnan},
  journal={arXiv preprint arXiv:2511.11698},
  year={2025}
}

@inproceedings{moiraimoe,
  title={Moirai-moe: Empowering time series foundation models with sparse mixture of experts},
  author={Liu, Xu and Liu, Juncheng and Woo, Gerald and Aksu, Taha and Liang, Yuxuan and Zimmermann, Roger and Liu, Chenghao and Savarese, Silvio and Xiong, Caiming and Sahoo, Doyen},
  booktitle={International Conference on Machine Learning},
  year={2025}
}

@article{chronos,
  title={Chronos: Learning the language of time series},
  author={Ansari, Abdul Fatir and Stella, Lorenzo and Turkmen, Caner and Zhang, Xiyuan and Mercado, Pedro and Shen, Huibin and Shchur, Oleksandr and Rangapuram, Syama Sundar and Arango, Sebastian Pineda and Kapoor, Shubham and others},
  journal={arXiv preprint arXiv:2403.07815},
  year={2024}
}

@article{chronos2,
  title={Chronos-2: From univariate to universal forecasting},
  author={Ansari, Abdul Fatir and Shchur, Oleksandr and K{\"u}ken, Jaris and Auer, Andreas and Han, Boran and Mercado, Pedro and Rangapuram, Syama Sundar and Shen, Huibin and Stella, Lorenzo and Zhang, Xiyuan and others},
  journal={arXiv preprint arXiv:2510.15821},
  year={2025}
}

@inproceedings{timesfm,
  title={A decoder-only foundation model for time-series forecasting},
  author={Das, Abhimanyu and Kong, Weihao and Sen, Rajat and Zhou, Yichen},
  booktitle={International Conference on Machine Learning},
  year={2024}
}

@article{toto2,
  title={Toto 2.0: Time Series Forecasting Enters the Scaling Era},
  author={Khwaja, Emaad and Lettieri, Chris and Woo, Gerald and Belouadah, Eden and Cenac, Marc and Jarry, Guillaume and Paquin, Enguerrand and Zhao, Xunyi and Zhukov, Viktoriya and Abou-Amal, Othmane and others},
  journal={arXiv preprint arXiv:2605.20119},
  year={2026}
}

@misc{gpt56sol,
  author={OpenAI},
  title={Previewing GPT-5.6 Sol: A Next-Generation Model},
  year={2026},
  howpublished={\url{https://openai.com/index/previewing-gpt-5-6-sol/}},
  note={Published June 26, 2026}
}

@misc{claudeopus48,
  author={Anthropic},
  title={Introducing Claude Opus 4.8},
  year={2026},
  howpublished={\url{https://www.anthropic.com/news/claude-opus-4-8}},
  note={Published May 28, 2026}
}

@misc{moiraiagent,
  author={Salesforce},
  title={MoiraiAgent: An Agentic Framework for Context-Aware Time-Series Forecasting},
  year={2026},
  howpublished={\url{https://www.salesforce.com/blog/moiraiagent/}},
  note={Published Jan 14, 2026}
}

@article{chatts,
  title={Chatts: Aligning time series with llms via synthetic data for enhanced understanding and reasoning},
  author={Xie, Zhe and Li, Zeyan and He, Xiao and Xu, Longlong and Wen, Xidao and Zhang, Tieying and Chen, Jianjun and Shi, Rui and Pei, Dan},
  journal={arXiv preprint arXiv:2412.03104},
  year={2024}
}

@inproceedings{chattime,
  title={Chattime: A unified multimodal time series foundation model bridging numerical and textual data},
  author={Wang, Chengsen and Qi, Qi and Wang, Jingyu and Sun, Haifeng and Zhuang, Zirui and Wu, Jinming and Zhang, Lei and Liao, Jianxin},
  booktitle={Proceedings of the AAAI Conference on Artificial Intelligence},
  volume={39},
  number={12},
  pages={12694--12702},
  year={2025}
}

@inproceedings{unitime,
  title={Unitime: A language-empowered unified model for cross-domain time series forecasting},
  author={Liu, Xu and Hu, Junfeng and Li, Yuan and Diao, Shizhe and Liang, Yuxuan and Hooi, Bryan and Zimmermann, Roger},
  booktitle={Proceedings of the ACM on Web Conference 2024},
  pages={4095--4106},
  year={2024}
}

@inproceedings{timeomni1,
  title={Timeomni-1: Incentivizing complex reasoning with time series in large language models},
  author={Guan, Tong and Meng, Zijie and Li, Dianqi and Wang, Shiyu and Yang, Chao-Han Huck and Wen, Qingsong and Liu, Zuozhu and Siniscalchi, Sabato and Jin, Ming and Pan, Shirui},
  booktitle={International Conference on Learning Representations},
  volume={2026},
  pages={152139--152170},
  year={2026}
}

@article{timeomnivl,
  title={Timeomni-vl: Unified models for time series understanding and generation},
  author={Guan, Tong and Pan, Sheng and Barthelemy, Johan and Li, Zhao and Cai, Yujun and Alippi, Cesare and Jin, Ming and Pan, Shirui},
  journal={arXiv preprint arXiv:2602.17149},
  year={2026}
}

@article{opentslm,
  title={Opentslm: Time-series language models for reasoning over multivariate medical text-and time-series data},
  author={Langer, Patrick and Kaar, Thomas and Rosenblattl, Max and Xu, Maxwell A and Chow, Winnie and Maritsch, Martin and Jakob, Robert and Wang, Ning and Liu, Juncheng and Verma, Aradhana and others},
  journal={arXiv preprint arXiv:2510.02410},
  year={2025}
}

@article{horai,
  title={Empowering Time Series Analysis with Large-Scale Multimodal Pretraining},
  author={Chen, Peng and Wang, Siyuan and Hu, Shiyan and Wu, Xingjian and Shu, Yang and Rao, Zhongwen and Wang, Meng and Li, Yijie and Yang, Bin and Guo, Chenjuan},
  journal={arXiv preprint arXiv:2602.05646},
  year={2026}
}

@article{thoth,
  title={Thoth: Mid-Training Bridges LLMs to Time Series Understanding},
  author={Lin, Jiafeng and Wang, Yuxuan and Wu, Jialong and Luo, Huakun and Pei, Zhongyi and Wang, Jianmin},
  journal={arXiv preprint arXiv:2603.01042},
  year={2026}
}

@inproceedings{llmstruggle,
  title={Language models still struggle to zero-shot reason about time series},
  author={Merrill, Mike A and Tan, Mingtian and Gupta, Vinayak and Hartvigsen, Thomas and Althoff, Tim},
  booktitle={Findings of the Association for Computational Linguistics: EMNLP 2024},
  pages={3512--3533},
  year={2024}
}

@article{llmuseful,
  title={Are language models actually useful for time series forecasting?},
  author={Tan, Mingtian and Merrill, Mike A and Gupta, Vinayak and Althoff, Tim and Hartvigsen, Thomas},
  journal={Advances in Neural Information Processing Systems},
  volume={37},
  pages={60162--60191},
  year={2024}
}

@article{gifteval,
  title={Gift-eval: A benchmark for general time series forecasting model evaluation},
  author={Aksu, Taha and Woo, Gerald and Liu, Juncheng and Liu, Xu and Liu, Chenghao and Savarese, Silvio and Xiong, Caiming and Sahoo, Doyen},
  journal={arXiv preprint arXiv:2410.10393},
  year={2024}
}

@article{mmtsbench,
  title={Mmts-bench: A comprehensive benchmark for time series understanding and reasoning},
  author={Yin, Yao and Xiao, Zhenyu and Li, Musheng and Liu, Yiwen and Nan, Sutong and He, Yiting and Wang, Ruiqi and Zhang, Zhenwei and Liao, Qingmin and Gu, Yuantao},
  journal={arXiv preprint arXiv:2602.08588},
  year={2026}
}

@article{timemmd,
  title={Time-mmd: Multi-domain multimodal dataset for time series analysis},
  author={Liu, Haoxin and Xu, Shangqing and Zhao, Zhiyuan and Kong, Lingkai and Kamarthi, Harshavardhan and Sasanur, Aditya B and Sharma, Megha and Cui, Jiaming and Wen, Qingsong and Zhang, Chao and others},
  journal={Advances in Neural Information Processing Systems},
  volume={37},
  pages={77888--77933},
  year={2024}
}

@article{mtbench,
  title={Mtbench: A multimodal time series benchmark for temporal reasoning and question answering},
  author={Chen, Jialin and Feng, Aosong and Zhao, Ziyu and Garza, Juan and Nurbek, Gaukhar and Qin, Cheng and Maatouk, Ali and Tassiulas, Leandros and Gao, Yifeng and Ying, Rex},
  journal={arXiv preprint arXiv:2503.16858},
  year={2025}
}

@inproceedings{timemqa,
  title={Time-mqa: Time series multi-task question answering with context enhancement},
  author={Kong, Yaxuan and Yang, Yiyuan and Hwang, Yoontae and Du, Wenjie and Zohren, Stefan and Wang, Zhangyang and Jin, Ming and Wen, Qingsong},
  booktitle={Proceedings of the 63rd Annual Meeting of the Association for Computational Linguistics (Volume 1: Long Papers)},
  pages={29736--29753},
  year={2025}
}

@inproceedings{cik,
  title={Context is key: A benchmark for forecasting with essential textual information},
  author={Williams, Andrew Robert and Ashok, Arjun and Marcotte, {\'E}tienne and Zantedeschi, Valentina and Subramanian, Jithendaraa and Riachi, Roland and Requeima, James and Lacoste, Alexandre and Rish, Irina and Chapados, Nicolas and others},
  booktitle={International Conference on Machine Learning},
  year={2025}
}

@inproceedings{temporalbench,
  title={Temporalbench: A benchmark for evaluating llm-based agents on contextual and event-informed time series tasks},
  author={Weng, Muyan and Cao, Defu and Yang, Wei and Sharma, Yashaswi and Liu, Yan},
  booktitle={Proceedings of the 32nd ACM SIGKDD Conference on Knowledge Discovery and Data Mining V. 2},
  pages={9997--10008},
  year={2026}
}

@article{motime,
  title={Motime: A dataset suite for multimodal time series forecasting},
  author={Zhou, Xin and Wang, Weiqing and Bald{\'a}n, Francisco J and Buntine, Wray and Bergmeir, Christoph},
  journal={arXiv preprint arXiv:2505.15072},
  year={2025}
}

@article{bedtime,
  title={Bedtime: A unified benchmark for automatically describing time series},
  author={Sen, Medhasweta and Gottesman, Zachary and Qiu, Jiaxing and Bruss, C Bayan and Nguyen, Nam and Hartvigsen, Tom},
  journal={arXiv preprint arXiv:2509.05215},
  year={2025}
}

@article{tsrbench,
  title={Tsrbench: A comprehensive multi-task multi-modal time series reasoning benchmark for generalist models},
  author={Yu, Fangxu and Guo, Xingang and Yuan, Lingzhi and Kang, Haoqiang and Zhao, Hongyu and Qin, Lianhui and Huang, Furong and Hu, Bin and Zhou, Tianyi},
  journal={arXiv preprint arXiv:2601.18744},
  year={2026}
}

@inproceedings{fintexts,
  title={Fintexts: Financial text-paired time-series dataset via semantic-based and multi-level pairing},
  author={Lee, Jaehoon and Park, Suhwan and Lim, Taeyoon and Lee, Seunghan and Seo, Jun and Kang, Dongwan and Choi, Hwanil and Kim, Minjae and Yoo, Sungdong and Lee, SoonYoung and others},
  booktitle={Proceedings of the 32nd ACM SIGKDD Conference on Knowledge Discovery and Data Mining V. 2},
  pages={9266--9277},
  year={2026}
}

@article{itstime,
  title={It's TIME: Towards the Next Generation of Time Series Forecasting Benchmarks},
  author={Qiao, Zhongzheng and Pan, Sheng and Wang, Anni and Zhukova, Viktoriya and Liu, Yong and Jiang, Xudong and Wen, Qingsong and Long, Mingsheng and Jin, Ming and Liu, Chenghao},
  journal={arXiv preprint arXiv:2602.12147},
  year={2026}
}

@article{fidelts,
  title={Fidel-TS: A High-Fidelity Multimodal Benchmark for Time Series Forecasting},
  author={Xu, Zhijian and Cai, Wanxu and Dai, Xilin and Deng, Zhaorong and Xu, Qiang},
  journal={arXiv preprint arXiv:2509.24789},
  year={2025}
}

@article{finmultitime,
  title={Finmultitime: A four-modal bilingual dataset for financial time-series analysis},
  author={Xu, Wenyan and Xiang, Dawei and Liu, Yue and Wang, Xiyu and Ma, Yanxiang and Zhang, Liang and Hu, Shu and Xu, Chang and Zhang, Jiaheng},
  journal={arXiv preprint arXiv:2506.05019},
  year={2025}
}
\bibliographystyle{iclr2027_conference}

\clearpage

\appendix

\section{Detailed Benchmark Statistics}
\label{app:benchmark-statistics}

\paragraph{Instance distribution.}
Table~\ref{tab:benchmark-statistics} summarizes the overall benchmark statistics. GIFT-Eval-CTX contains 10,000 univariate forecasting instances drawn from the held-out test regions of 20 GIFT-Eval source datasets. A source dataset may appear at multiple sampling frequencies, giving 38 distinct dataset--frequency identifiers. Combining these identifiers with the available short-, medium-, and long-horizon settings produces 78 dataset--frequency--term configurations. Table~\ref{tab:source-statistics} provides a source-level breakdown, where \textit{IDs} denotes the number of dataset--frequency identifiers and \textit{Groups} denotes the number of dataset--frequency--term configurations. Table~\ref{tab:frequency-horizon-statistics} reports the number of instances at each sampling frequency and forecast horizon.

\begin{table}[h]
    \centering
    \caption{Summary statistics for GIFT-Eval-CTX.}
    \vspace{0.5em}
    \label{tab:benchmark-statistics}
    \begin{tabular}{lr}
        \toprule
        Statistic & Value \\
        \midrule
        Instances & 10,000 \\
        Source datasets & 20 \\
        Dataset--frequency identifiers & 38 \\
        Dataset--frequency--term configurations & 78 \\
        Sampling frequencies & 6 \\
        Distinct forecast horizons & 7 \\
        Horizon range & 30--900 \\
        Short / medium / long instances & 6,559 / 2,135 / 1,306 \\
        Micro effects & 23,476 \\
        Mean Micro effects per instance & 2.35 \\
        \bottomrule
    \end{tabular}
\end{table}

\begin{table*}[h]
    \centering
    \caption{Source-level composition of GIFT-Eval-CTX.}
    \vspace{0.5em}
    \label{tab:source-statistics}
    \resizebox{0.9\textwidth}{!}{%
    \begin{tabular}{lrrrrrr}
        \toprule
        Source dataset & IDs & Groups & Short & Medium & Long & Total \\
        \midrule
        LOOP\_SEATTLE & 3 & 7 & 1,332 & 685 & 306 & 2,323 \\
        M\_DENSE & 2 & 4 & 61 & 5 & 4 & 70 \\
        SZ\_TAXI & 2 & 4 & 279 & 86 & 73 & 438 \\
        bitbrains\_fast\_storage & 2 & 4 & 799 & 257 & 112 & 1,168 \\
        bitbrains\_rnd & 2 & 4 & 681 & 100 & 97 & 878 \\
        bizitobs\_application & 1 & 2 & 1 & 1 & 0 & 2 \\
        bizitobs\_l2c & 2 & 5 & 19 & 12 & 12 & 43 \\
        bizitobs\_service & 1 & 3 & 46 & 18 & 13 & 77 \\
        covid\_deaths & 1 & 1 & 23 & 0 & 0 & 23 \\
        electricity & 3 & 7 & 1,184 & 707 & 502 & 2,393 \\
        ett1 & 3 & 7 & 28 & 21 & 20 & 69 \\
        ett2 & 3 & 7 & 31 & 25 & 23 & 79 \\
        hierarchical\_sales & 1 & 1 & 97 & 0 & 0 & 97 \\
        jena\_weather & 3 & 7 & 56 & 33 & 33 & 122 \\
        kdd\_cup\_2018\_with\_missing & 2 & 4 & 572 & 18 & 11 & 601 \\
        m4\_hourly & 1 & 1 & 35 & 0 & 0 & 35 \\
        restaurant & 1 & 1 & 130 & 0 & 0 & 130 \\
        saugeenday & 1 & 1 & 1 & 0 & 0 & 1 \\
        solar & 3 & 7 & 641 & 167 & 100 & 908 \\
        temperature\_rain\_with\_missing & 1 & 1 & 543 & 0 & 0 & 543 \\
        \midrule
        Total & 38 & 78 & 6,559 & 2,135 & 1,306 & 10,000 \\
        \bottomrule
    \end{tabular}%
    }
\end{table*}

\paragraph{Structured-effect distribution.}
Each benchmark instance contains one Macro annotation and at least one Micro effect. In total, the benchmark contains 23,476 Micro effects, with a median of two and a maximum of 15 effects per instance. Excursions constitute 19,134 effects, seasonal changes 3,117, and mean shifts 1,225. Table~\ref{tab:micro-subtype-statistics} provides the complete subtype breakdown.

\begin{table}[h]
    \centering
    \caption{Distribution of sampling frequency and forecast horizon.}
    \vspace{0.5em}
    \label{tab:frequency-horizon-statistics}
    \resizebox{0.5\columnwidth}{!}{%
    \begin{tabular}{lrlr}
        \toprule
        Frequency & Instances & Horizon & Instances \\
        \midrule
        10S & 79    & 30  & 1,791 \\
        5T  & 2,724 & 48  & 4,721 \\
        10T & 543   & 60  & 47 \\
        15T & 1,597 & 480 & 2,116 \\
        H   & 3,266 & 600 & 19 \\
        D   & 1,791 & 720 & 1,293 \\
            &       & 900 & 13 \\
        \bottomrule
    \end{tabular}
    }
    \vspace{-1em}
\end{table}

\begin{table}[h]
    \centering
    \caption{Distribution of Micro families and subtypes.}
    \vspace{0.5em}
    \label{tab:micro-subtype-statistics}
    \resizebox{0.45\columnwidth}{!}{%
    \begin{tabular}{llr}
        \toprule
        Family & Subtype & Count \\
        \midrule
        \multirow{3}{*}{Excursion}
            & Impulse & 10,625 \\
            & Sustained & 6,750 \\
            & Recurrent & 1,759 \\
        \midrule
        Mean shift & -- & 1,225 \\
        \midrule
        \multirow{5}{*}{Seasonal change}
            & Amplitude & 2,391 \\
            & Phase & 29 \\
            & Period & 24 \\
            & Onset & 211 \\
            & Disappearance & 462 \\
        \bottomrule
    \end{tabular}
    }
\end{table}

\section{Complete Structured Effects Schema}
\label{app:complete-schema}

\subsection{Macro Schema}
\label{app:macro-schema}

Macro describes the complete forecast horizon. Table~\ref{tab:macro-schema} lists its fields and constraints.

\begin{table*}[h]
    \centering
    \caption{Complete Macro schema.}
    \vspace{0.5em}
    \label{tab:macro-schema}
    \small
    \resizebox{\textwidth}{!}{%
    \begin{tabular}{p{0.27\textwidth}p{0.27\textwidth}p{0.43\textwidth}}
        \toprule
        Field & Type and allowed values & Definition \\
        \midrule
        \texttt{trend}
            & List of one or two segment objects
            & Chronological decomposition of the horizon-wide center-line movement. Segment
              durations are positive and sum to one. \\
        \texttt{trend.direction}
            & \texttt{up}, \texttt{down}, or \texttt{flat}
            & Direction of the center line within the segment. Flat means there is no meaningful overall
            rise or fall. \\
        \texttt{trend.shape}
            & \texttt{linear}, \texttt{curved}, \texttt{accelerating\_curved},
              \texttt{decelerating\_curved}, \texttt{step}, or null
            & Shape of a non-flat trend segment. The value is null when direction is
              \texttt{flat}. \\
        \texttt{trend.duration}
            & Positive real number in $(0,1]$
            & Fraction of the forecast horizon occupied by the segment. \\
        \texttt{seasonality}
            & Object or null
            & Dominant oscillatory structure of the future. It is null when no meaningful
              oscillation is present. \\
        \texttt{seasonality.period}
            & Integer greater than one, or null
            & Dominant period in time steps. A null period denotes an irregular oscillation. \\
        \texttt{seasonality.waveform}
            & \texttt{follow\_history}, \texttt{sinusoidal}, \texttt{sawtooth}, or
              \texttt{irregular}
            & Qualitative shape of the dominant oscillation. \\
        \texttt{level\_range}
            & Two finite real values $[\ell_{\min},\ell_{\max}]$
            & Minimum and maximum future values in history-normalized coordinates. The field is
              computed deterministically from the complete future. \\
        \bottomrule
    \end{tabular}
    }
\end{table*}

\subsection{Micro Schema}
\label{app:micro-schema}

A Micro entry describes one interval-specific mechanism. Multiple entries may overlap when they represent different compatible mechanisms. Each entry contains the following fields:

\begin{itemize}
    \item \texttt{family}: the top-level mechanism: \texttt{excursion}, \texttt{mean\_shift}, or \texttt{seasonal\_change}.
    \item \texttt{subtype}: the specific form of the mechanism. Its valid values depend on the family.
    \item \texttt{direction}: the direction of the change relative to the locally reasonable background.
    \item \texttt{amplitude\_trend}: how component or cycle amplitudes evolve within the interval: \texttt{increase}, \texttt{decrease}, \texttt{stable}, \texttt{variable}, or null.
    \item \texttt{envelope\_change}: how the overall envelope of a bidirectional recurrent pattern evolves: \texttt{expanding}, \texttt{contracting}, \texttt{stable}, \texttt{variable}, or null.
    \item \texttt{start} and \texttt{end}: zero-based, inclusive boundaries within the future horizon.
    \item \texttt{intensity}: a deterministic, family-specific numerical measurement.
\end{itemize}

\paragraph{Shared normalization.}
Let $x_{\min}$ and $x_{\max}$ be the minimum and maximum finite history values, and let $R_x=x_{\max}-x_{\min}$. A normalized level $v$ is represented as $\ell(v)=100(v-x_{\min})/R_x$, and the inverse transformation is $v=x_{\min}+\ell(v)R_x/100$. Values below 0 or above 100 are valid when the future lies outside the historical range. Macro \texttt{level\_range} is $[\ell(\min_t y_t),\ell(\max_t y_t)]$. Micro level intensities use the same coordinates.

\paragraph{Excursion.}
An excursion is a temporary departure from the locally reasonable background followed by a meaningful recovery.

\begin{itemize}
    \item \textbf{Subtype:} \texttt{impulse} is one isolated, sharply concentrated peak or trough; \texttt{sustained} is one continuous deviation that persists locally before recovering; and \texttt{recurrent} represents multiple separated events with quiet intervals between them rather than a stable seasonal cycle.
    \item \textbf{Direction:} \texttt{impulse} and \texttt{sustained} are \texttt{up} or \texttt{down}; \texttt{recurrent} is \texttt{up}, \texttt{down}, or \texttt{bidirectional}.
    \item \textbf{Amplitude fields:} for a unidirectional recurrent excursion, \texttt{amplitude\_trend} is \texttt{increase}, \texttt{decrease}, \texttt{stable}, or \texttt{variable}, and \texttt{envelope\_change} is null. For a bidirectional recurrent excursion, \texttt{envelope\_change} is \texttt{expanding}, \texttt{contracting}, \texttt{stable}, or \texttt{variable}, and \texttt{amplitude\_trend} is null. Both fields are null for \texttt{impulse} and \texttt{sustained}.
    \item \textbf{Boundaries:} when available, \texttt{start} includes the point immediately before the first affected point so that the onset is visible. \texttt{end} includes the first clear recovery point when recovery occurs within the horizon.
    \item \textbf{Intensity:} for \texttt{impulse} and \texttt{sustained}, intensity is the interval maximum for \texttt{up}, or the interval minimum for \texttt{down}, expressed in history-normalized level coordinates. Thus it is $\ell(\max_{t\in[a,b]}y_t)$ for \texttt{up} and $\ell(\min_{t\in[a,b]}y_t)$ for \texttt{down}. For \texttt{recurrent}, intensity is a representative component value expressed in the same history-normalized level coordinates. For an upward recurrent excursion, we use the peak value of each component; for a downward recurrent excursion, we use the trough value of each component. When \texttt{amplitude\_trend} is \texttt{increase} or \texttt{decrease}, intensity is the maximum peak for \texttt{up} or the minimum trough for \texttt{down}. When it is \texttt{stable} or \texttt{variable}, intensity is the median component value. For a bidirectional recurrent excursion, the up-component peaks and down-component troughs are combined: intensity is their maximum for an \texttt{expanding} envelope, their minimum for a \texttt{contracting} envelope, and their median for a \texttt{stable} or \texttt{variable} envelope.
\end{itemize}

\paragraph{Mean shift.}
A mean shift is a transition from the nearest preceding stable level into a persistent new level. Its \texttt{subtype} is null, and its \texttt{direction} is \texttt{up} when the settled new level is above the preceding stable level and \texttt{down} when it is below. Both \texttt{amplitude\_trend} and \texttt{envelope\_change} are null. Intensity records the settled new level as $100(\widetilde{y}-x_{\min})/R_x$, where $\widetilde{y}$ is the median settled level.

\paragraph{Seasonal change.}
A seasonal change describes a change in a periodic pattern relative to the nearest stable periodic region immediately preceding \texttt{start}. There are five subtypes in this family: \texttt{amplitude}, \texttt{phase}, \texttt{period}, \texttt{onset}, and \texttt{disappearance}. The subtype and its corresponding direction meaning is:

\begin{itemize}
    \item \texttt{amplitude}: \texttt{increase} direction means that cycle amplitudes are larger than in the preceding stable region, and \texttt{decrease} means that they are smaller.
    \item \texttt{phase}: \texttt{advance} direction means that the recurring pattern occurs earlier within each cycle, and \texttt{delay} means that it occurs later.
    \item \texttt{period}: \texttt{increase} direction means that cycles become longer, and \texttt{decrease} means that they become shorter.
    \item \texttt{onset}: a periodic pattern appears after being absent from the preceding stable region; \texttt{direction} is null.
    \item \texttt{disappearance}: a periodic pattern present in the preceding stable region disappears; \texttt{direction} is null.
\end{itemize}

For \texttt{amplitude} and \texttt{onset} subtypes, the field of \texttt{amplitude\_trend} describes how cycle amplitudes evolve inside the labeled interval. It is required when the interval contains at least two complete cycles and is null when it contains exactly one. For example, \texttt{seasonal\_change/amplitude/decrease} with \texttt{amplitude\_trend=increase} means that amplitudes are smaller than in the preceding stable region but increase within the labeled interval. For \texttt{phase}, \texttt{period}, and \texttt{disappearance}, the field of \texttt{amplitude\_trend} is null, and the field of \texttt{envelope\_change} is null for every seasonal-change subtype.

Intensity is defined separately for each seasonal-change subtype:
\begin{itemize}
    \item For \texttt{amplitude} or \texttt{onset}, intensity encodes a representative peak-to-trough cycle amplitude $A^\star$, with $A^\star=\texttt{intensity}\,R_x/100$. If the interval contains one complete cycle, it is that cycle's amplitude. With multiple complete cycles, it is the maximum amplitude when \texttt{amplitude\_trend} is \texttt{increase} or \texttt{decrease}, and the median when it is \texttt{stable} or \texttt{variable}.
    \item For \texttt{phase}, intensity is $100\Delta/P_{\mathrm{ref}}$, where $\Delta$ is the shortest circular phase shift and $P_{\mathrm{ref}}$ is the preceding period. The direction specifies whether the waveform advances or is delayed.
    \item For \texttt{period}, intensity is $100|P_{\mathrm{new}}-P_{\mathrm{ref}}|/P_{\mathrm{ref}}$. The direction specifies whether the period increases or decreases.
    \item For \texttt{disappearance}, intensity is null.
\end{itemize}

\subsection{Micro Effect Examples}
\label{app:micro-effect-examples}

\begin{figure*}[t]
    \centering
    \includegraphics[width=\textwidth]{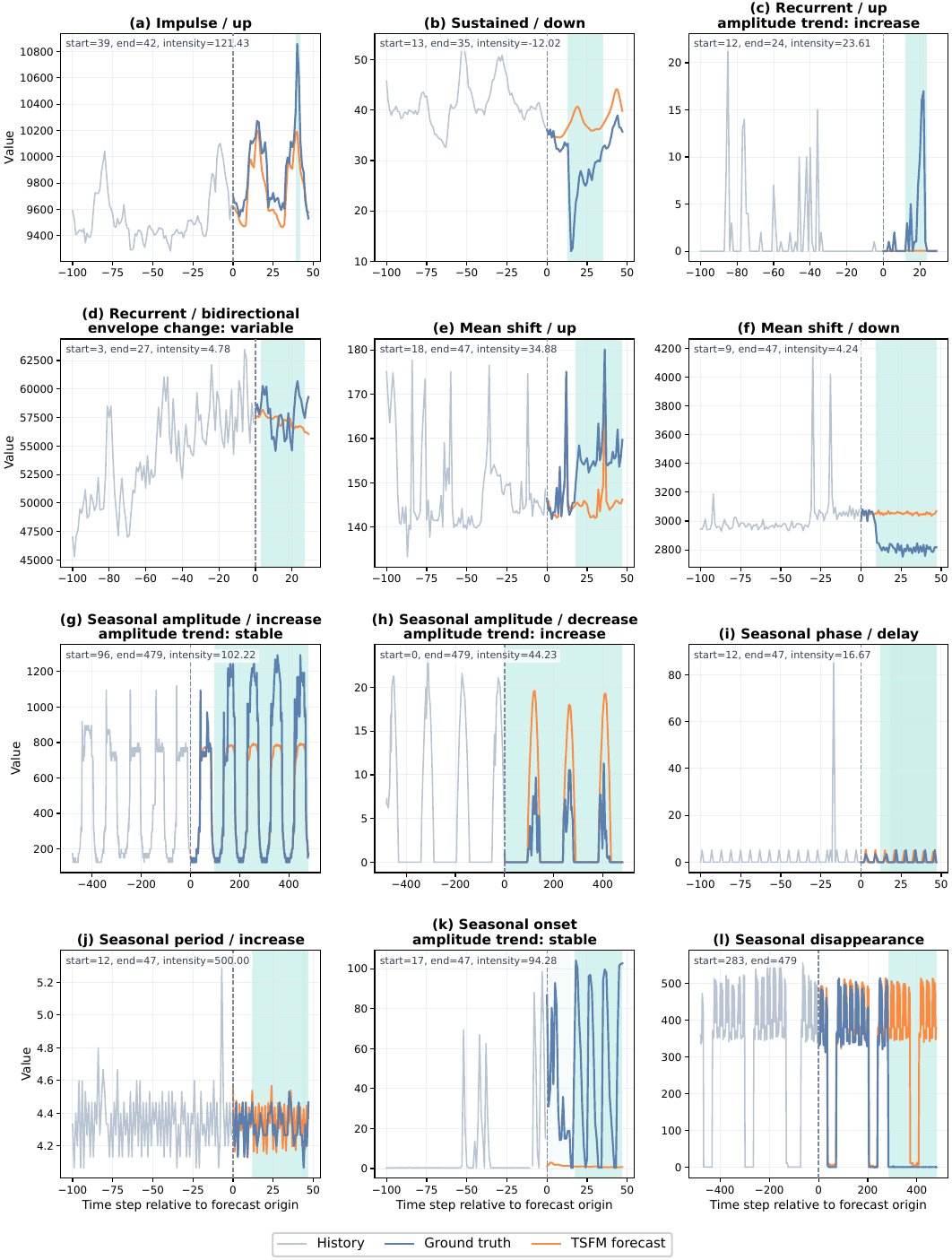}
    \caption{Representative examples of the Micro-effect taxonomy. Shaded regions indicate the inclusive intervals described by the effects.}
    \label{fig:effect-examples}
\end{figure*}

Figure~\ref{fig:effect-examples} visualizes representative examples from the Micro-effect taxonomy. Panels (a)--(d) show the three excursion subtypes: impulse, sustained, and recurrent. Panel (c) contains upward recurrent excursions whose amplitudes increase, while panel (d) contains bidirectional recurrent excursions with a variable envelope. Panels (e) and (f) show upward and downward mean shifts. Panels (g)--(l) cover the seasonal-change subtypes. Panels (g) and (h) distinguish change relative to the preceding periodic region from amplitude evolution within the labeled interval: panel (g) has increased and stable amplitudes, whereas panel (h) has decreased amplitudes that subsequently increase. Panels (i)--(l) show a phase delay, period increase, seasonal onset, and seasonal disappearance, respectively.

\section{Benchmark Construction}
\label{app:construction-generation}

\subsection{Time Series Corpus Selection}

We begin with GIFT-Eval test windows whose forecast horizons fall between 24 and 1,000 steps. Multivariate collections are expanded into independent univariate targets. No candidate is cut from a training or validation region. For a horizon of length $H$, the most recent $10H$, $5H$, and $3H$ historical observations are retained for short-, medium-, and long-horizon configurations, respectively. A candidate is discarded if its future contains a missing value, more than 20\% of its history is missing, or the finite history or future is effectively constant, defined by a standard deviation below $10^{-8}$.

Each candidate is forecast independently by top-performing models on GIFT-Eval. We use the median prediction from each model and compute $\operatorname{rMAE}(\hat{\mathbf y},\mathbf y)= \frac{H^{-1}\sum_{t=1}^{H}|\hat y_t-y_t|}{\max_t y_t-\min_t y_t}$. A candidate is retained only if all models obtain $\operatorname{rMAE}\geq0.10$. Thus, each model's mean absolute error must be at least 10\% of the future value range. This deliberately removes windows that are too easy for the TSFMs. After retaining a window, we skip enough subsequent windows from the same univariate target to ensure that the complete history--future spans of retained instances do not overlap. This prevents similar windows from entering the subsequent annotation process. To prevent a few large sources from dominating the benchmark, medium- and long-horizon configurations are capped at approximately 600 and 300 retained windows, respectively. The largest short-horizon configurations receive source-specific caps, and capped sources are sampled with a fixed seed of 42.

\subsection{Structured Effects Generation}

\paragraph{Macro generation.}
The Macro Agent receives the complete history and future ground truth. It can inspect deterministic measurements of trend, change points, and periodic structure before proposing a complete Macro object. The Macro validator enforces the schema, checks that trend durations sum to one, and replaces the proposed \texttt{level\_range} with the value computed directly from the future. Trend and seasonality measurements are returned as evidence for reconsideration when they disagree with a proposal.

\paragraph{Micro generation.}
For each retained instance, we construct a TSFM reference forecast by pointwise averaging the predictions of top-performing TSFMs. We compare this reference with the future ground truth and use the forecast residuals for region selection. Regions with large residuals indicate where the TSFMs forecast poorly and where future context may be most useful. The Micro Agent then uses an overview tool to locate these regions and inspection tools to retrieve their exact raw values and surrounding context. Based on this evidence, the Agent identifies the underlying mechanism and proposes its family, subtype, direction, and complete interval. Mean shifts and seasonal changes additionally use the nearest stable region immediately preceding the label. Reference coordinates may lie in history, cross the history--future boundary, or lie in the future. Seasonal comparisons use complete cycles rather than isolated peaks.

\paragraph{Deterministic Micro validation.}
Each proposal is checked against the raw time series by a family-specific deterministic validator. The validator verifies that the proposed mechanism and boundaries are supported by numerical evidence. It also requires the labeled region to coincide with a substantial error in the TSFM reference forecast. Proposals that lack sufficient evidence are rejected. For accepted proposals, the validator computes the family-specific intensity defined in Appendix~\ref{app:micro-schema}; otherwise, it returns numerical evidence that allows the Agent to inspect the region again and revise its proposal.

\paragraph{Blind-first joint review.}
After both discovery branches finish, the Joint Critic Agent first inspects the raw history and future while all proposals remain hidden. It then views the Macro and Micro candidates, their rationales, and validator evidence. For each Micro candidate, the Critic may support, exclude, or revise the proposal; revised candidates pass through the same Micro validator as above. Only supported Micro effects enter the final annotation. This ordering reduces anchoring on discovery outputs and ensures that the final decision is based on an independent reading of the raw series. We additionally assess annotation quality by manually reviewing a random subset of annotated samples.

\section{Experimental Details}
\label{app:experimental-details}

\subsection{Model Configurations}

Table~\ref{tab:model-configurations} summarizes the method configurations. All TSFMs operate in the univariate setting and return their median point forecast. The frontier LLMs are accessed through API calls, allowing up to 16,000 output tokens and using medium reasoning effort. MoiraiAgent combines the API-accessed GPT-5.6 Sol with an agent harness built around the public Moirai-2 checkpoint. Concretely, MoiraiAgent reads the numerical history from a session-local CSV file. It first calls the Moirai-2 forecasting tool to obtain an exact-length numerical trajectory, then uses the supplied effects and code-execution tools to inspect and modify that trajectory. The final forecast must be written as a finite, exact-length JSON array. Each run permits at most ten Agent steps and 16,000 output tokens.

\begin{table*}[h]
    \centering
    \caption{Model configurations used in the experiments.}
    \vspace{0.5em}
    \label{tab:model-configurations}
    \resizebox{\textwidth}{!}{%
    \begin{tabular}{lll}
        \toprule
        Method & Model or checkpoint & Forecasting interface \\
        \midrule
        Seasonal naive & -- & Repeats the last observed seasonal cycle \\
        Chronos-2 & \texttt{amazon/chronos-2} & Direct median forecast \\
        Toto-2 & \texttt{Datadog/Toto-2.0-313m} & Direct median forecast \\
        Moirai-2 & \texttt{Salesforce/moirai-2.0-R-small} & Direct median forecast \\
        TimesFM-3 & \texttt{google/timesfm-3.0-pytorch} & Direct median forecast \\
        GPT-5.6 Sol & GPT-5.6 Sol & Direct numerical generation \\
        Claude Opus 4.8 & Claude Opus 4.8 & Direct numerical generation \\
        MoiraiAgent & GPT-5.6 Sol + Moirai-2 & Forecast and code-execution tools \\
        \bottomrule
    \end{tabular}
    }
\end{table*}

\subsection{LLM Prompts}

We use separate system prompts for the history-only and history-with-effects conditions. The prompt construction is given below.

\paragraph{History only.}
The system message is:
\begin{quote}\small
You are a time series forecasting expert. You will be given historical values of a target variable and the forecast horizon. Analyze the history to identify its trend, seasonality, level, and noise, then generate exact numerical forecast values for the specified horizon. The forecast length must equal the specified horizon. Return a JSON object with exactly one key, \texttt{\{"forecast": [<H floats>]\}}, and no other text.
\end{quote}
The user message is instantiated as:
\begin{quote}\small
Forecast horizon: \texttt{\{H\}} steps\\ Historical values: $[x_1, \ldots, x_T]$\\[0.4em] Respond with only the JSON object \texttt{\{"forecast": [...]\}}. The forecast array must contain exactly \texttt{\{H\}} numerical values---no more and no fewer. Count and verify the values before finishing. No analysis or prose.
\end{quote}
We repeat the exact-length requirement at the end of the user message because initial experiments showed that direct LLMs frequently failed to return the requested number of forecast values.

\paragraph{History + effects.}
The system message is:
\begin{quote}\small
You are a time series forecasting expert. You will be given historical values of a target variable, the forecast horizon, and structured effects describing how the future horizon will behave. The structured effects contain two complementary parts: Macro describes the overall structure of the complete future horizon, while Micro describes interval-specific mechanisms. Treat the structured effects as reliable guidance and use them together with the historical values to produce the future numerical values.

\textit{[Macro schema and definitions]}\\
\textit{[Micro schema and definitions]}

Study the history to understand the scale, recent level, and the shape and amplitude of any oscillation so that the forecast connects naturally to the history. Produce a forecast that makes full use of both parts of the structured effects: Macro describes the overall behavior of the future horizon, and Micro describes interval-specific mechanisms. The forecast length must equal the specified horizon. Return a JSON object with exactly one key, \texttt{\{"forecast": [<H floats>]\}}, with no analysis, prose, Markdown, or code fences.
\end{quote}
The corresponding user message is:
\begin{quote}\small
Forecast horizon: \texttt{\{H\}} steps\\ Historical values: $[x_1, \ldots, x_T]$\\[0.4em] Structured effects (describing future behavior):\\ \texttt{\{"macro": \{<Macro object>\}, "micro": [<Micro objects>]\}}\\[0.4em] For this sample, the Macro \texttt{level\_range} corresponds to the inclusive raw-value range \texttt{[\{L\}, \{U\}]} over the forecast horizon. Every forecast value must remain within this range.\\[0.4em] Respond with only the JSON object \texttt{\{"forecast": [...]\}}. The forecast array must contain exactly \texttt{\{H\}} numerical values---no more and no fewer. Count and verify the values before finishing. No analysis or prose.
\end{quote}
We additionally provide \texttt{level\_range} in raw-value units because direct LLMs sometimes miscalculate the conversion from normalized coordinates, resulting in disproportionately large forecast errors.

\subsection{Metrics and Aggregation}

Let $m$ be the seasonal period associated with the sampling frequency. For a sample with history $\mathbf x_{1:T}$, future $\mathbf y_{1:H}$, and forecast $\hat{\mathbf y}_{1:H}$, we compute \[ \operatorname{MASE} =\frac{H^{-1}\sum_{t=1}^{H}|y_t-\hat y_t|} {(T-m)^{-1}\sum_{t=m+1}^{T}|x_t-x_{t-m}|}. \] If $m\geq T$, the denominator falls back to the one-step naive error. Seasonal periods follow the GluonTS frequency mapping. The seasonal-naive forecast repeats the last complete seasonal cycle, or the historical mean when the history is shorter than one cycle.

Aggregation follows GIFT-Eval. For each of the 78 dataset--frequency--term groups, we first take the arithmetic mean of per-sample MASE for each method. We divide this group mean by the corresponding seasonal-naive group mean, then take the geometric mean of the resulting ratios.

\section{Additional Related Work}
\subsection{Time Series Foundation Models and Multimodal Models}

Time series foundation models have achieved strong zero-shot forecasting performance~\citep{moirai,moiraimoe,chronos}, but their standard interfaces operate primarily on numerical values and cannot directly incorporate multimodal context. Multimodal time series methods address this limitation through several modeling strategies. HORAI uses large-scale multimodal pretraining to incorporate endogenous and exogenous information~\citep{horai}. ChatTS, ChatTime, and OpenTSLM adapt LLMs to represent and reason over time series alongside language~\citep{chatts,chattime,opentslm}. More recent systems, including Thoth, TimeOmni-1, and TimeOmni-VL, extend this direction toward broader time series reasoning and unified understanding and generation~\citep{thoth,timeomni1,timeomnivl}.

Many of these LLM-based approaches build on small open-weight LLMs to keep training feasible. These base models generally offer less world knowledge and weaker reasoning and agentic capabilities than frontier LLMs. Systems built on them may therefore be insufficient for real-world contextual forecasting, where relevant information can be broad, heterogeneous, domain-specific, and semantically complex. GIFT-Eval-CTX instead evaluates frontier LLMs and agentic systems built around them.

\end{document}